\documentclass[sn-basic]{sn-jnl}

\usepackage{graphicx}%
\usepackage{multirow}%
\usepackage{amsmath,amssymb,amsfonts}%
\usepackage{amsthm}%
\usepackage{mathrsfs}%
\usepackage[title]{appendix}%
\usepackage{xcolor}%
\usepackage{textcomp}%
\usepackage{manyfoot}%
\usepackage{booktabs}%
\usepackage{algorithm}%
\usepackage{algorithmicx}%
\usepackage{algpseudocode}%
\usepackage{listings}%
\usepackage{braket}
\usepackage{bm}
\usepackage{qcircuit}
\usepackage{here}
\usepackage{hhline}

\theoremstyle{thmstyleone}%
\theoremstyle{thmstyletwo}%

\theoremstyle{thmstylethree}%

\newcommand*{\figref}[1]{\figurename~\ref{#1}}
\newcommand*{\tblref}[1]{\tablename~\ref{#1}}

\begin{document}

\title[The cross-sectional stock return predictions via quantum neural network and tensor network]{The cross-sectional stock return predictions via quantum neural network and tensor network}


\author*[1]{\fnm{Nozomu} \sur{Kobayashi}}\email{nozomu.kobayashi@nomura.com}

\author[1]{\fnm{Yoshiyuki} \sur{Suimon}}

\author[2]{\fnm{Koichi} \sur{Miyamoto}}
\author[3,2]{\fnm{Kosuke} \sur{Mitarai}}

\affil*[1]{\orgdiv{Data Science Department}, \orgname{Nomura Securities Co., Ltd.}, \orgaddress{\city{Tokyo}, \country{Japan}}}
\affil[2]{\orgdiv{Center for Quantum Information and Quantum Biology}, \orgname{Osaka University}, \orgaddress{\city{Osaka}, \country{Japan}}}
\affil[3]{\orgdiv{Graduate School of Engineering Science}, \orgname{Osaka University}, \orgaddress{\city{Osaka}, \country{Japan}}}

\abstract{
    In this paper, we investigate the application of quantum and quantum-inspired machine learning algorithms to stock return predictions.
    Specifically, we evaluate the performance of quantum neural network, an algorithm suited for noisy intermediate-scale quantum computers, and tensor network, a quantum-inspired machine learning algorithm, against classical models such as linear regression and neural networks. 
    To evaluate their abilities, we construct portfolios based on their predictions and measure investment performances.
    The empirical study on the Japanese stock market shows the tensor network model achieves superior performance compared to classical benchmark models, including linear and neural network models. Though the quantum neural network model attains a lowered risk-adjusted excess return than the classical neural network models over the whole period, both the quantum neural network and tensor network models have superior performances in the latest market environment, which suggests the capability of the model's capturing non-linearity between input features.
}

\keywords{Quantum Machine Learning, Tensor Network, Finance, Stock Return Prediction}

\maketitle

\section{Introduction}

The arrival of real quantum computers and experiments that show the quantum supremacy \cite{arute2019quantum,madsen2022quantum} make it more realistic that the new computational paradigm will come by virtue of quantum computing. 
It is true that we are currently at the era of NISQ (noisy intermidiate-scale quantum computer) \cite{Preskill2018quantumcomputingin} and must implement the quantum error correction for a full picture of such a new paradigm, but rapid progress of quantum technologies already open a new window to the research in various fields, such as quantum chemistry, optimization, machine learning, and finance.
It is therefore worth looking for a practical application of quantum computers even in the NISQ era. 
The framework of variational quantum algorithms (VQAs) \cite{cerezo2021varational} is thought to be an effective approach towards the goal.
It has been applied, for instance, to solve machine learning problems \cite{mitarai2018quantum,schuld2019evaluating}.

Machine learning techniques developed within the framework of VQAs are essentially equivalent to the ones using tensor networks \cite{huggins2019towards,stoudenmire2016supervised,stoudenmire2018learning}, which is originally invented as a tool to simulate quantum physics in classical computers \cite{fannes1992finitely,white1992density}.
Its ability to utilize an exponentially large tensor into a factorized series of smaller tensors has also allowed the machine learning community to successfully solve various machine learning problems \cite{novikov2016exponential,stoudenmire2016supervised,stoudenmire2018learning}.
It can consequently be considered a quantum-inspired machine learning algorithm.

Given these growing interests of quantum and quantum-inspired machine learning algorithms, it is important to study their applicability on the real-world problems, which are, however, less known so far partly due to the current limitation of computational resource for quantum computers and their simulators.
In this work, to address the above issue, we consider a real-world financial problem, namely the prediction of stock returns, employing quantum and quantum-inspired machine learning algorithms.

Stock return prediction has been a principal problem in finance. 
Ever since the work by Fama and French \cite{eugene1992cross}, who have provided the empirical evidence that the notion of so-called factors is effective in return explainability, significant efforts have been made to find unseen factors that have predictable powers for stock returns. Among practical investors, multi-factor models, which is a linear regression of stock returns by a set of factors, are commonly used thanks to their simplicity and interpretability, though they lack expressibility due to the absence of interaction terms between factors.
As it is, machine learning has been becoming an alternative to them. 
Various studies, \cite{abe2018deep,chinco2019sparse,dixon2020deep,10.1093/rfs/hhaa009,gu2021autoencoder} to name but a few, are conducted on stock return predictions with machine learning methods, which can capture non-linearity in contrast to multi-factor models.

Our interest here is to test whether the quantum or quantum-inspired techniques can be applied to predict stock returns and also have a competitive advantage over classical machine learning algorithms in that task.
To this end, using a set of stocks in the Japanese stock market, we conduct portfolio backtesting over 10 years based on stock return predictions by quantum neural network, tensor network, standard linear regression, and neural network and compare their performances.
As a result, we find that the tensor network model outperforms the other models, while the quantum neural network model is inferior to the neural network model in the whole backtesting period.
We also observe that in the latest market environment, the quantum neural network model has the better performance than the neural network model, which might be related to the overfitting problem.
This experiment provides the implication that quantum neural network and tensor network may be able to learn non-linear and interaction effects among features, and they have potential to use in return predictions beyond the conventional models.


This paper is organized as follows.
In Section \ref{sec:method}, we explain the definition of our problem, the stock return predictions, and then describe both classical and quantum machine learning algorithms we use in our analysis.
Section \ref{sec:experiment} presents the methodology to conduct our backtesting experiment and then shows its results, using quantitative metrics that are often used to evaluate an investment performance.
Finally, in Section \ref{sec:conclusion} we conclude our analysis and discuss some future directions for further research.

\section{Methodology}
\label{sec:method}
This section collects all the ingredients we use in our analysis. 
First, we set up the definition of our objective as stock return prediction by cross-sectional analysis, which is based on comparing each stock to others at a point in time, and describe general methodology to tackle the problem.
Then, we explain classical models for return predictions, namely the linear regression and the neural network models.
Both models are used as benchmark models against quantum ones in our experiment, according to the following reasons. 
The linear regression model is one of the most traditional models as well as widely employed both by academicians and practitioners in finance.
The neural network model serves as a classical counterpart of quantum models, not to mention that it shows superior investing performance over the linear model thanks to its flexibility and expressibility.
After that, we introduce quantum circuit learning, which is one realization of quantum neural network, and tensor network algorithms in our framework.
Finally, we describe the optimization procedure for these machine learning models.
 
\subsection{Problem Definition}
The objective of this work is to predict stock returns over cross-section.
Before formulating our problem, let us clarify some notations.

Suppose there are trading dates indexed as $0 \leq t \leq T$, and at each trading date $t$, we have $N_t$ stocks available to invest.
We call such a whole set of stocks as stock universe and denote $U_t$. 
Remark that the frequency of trading periods depends on our purpose and data availability, by which we adjust the frequency for monthly, weekly, daily, and so on.
We describe most generic situation that stock universe varies over time.
Stocks are indexed as $i= 1, \cdots , N_t$, and price of $i$-th stock at time $t$ is denoted as $p_{i,t}$.
The return of $i$-th stock from $t$ to $t+1$ is then calculated as  
\begin{align}
    r_{i,t+1} = \frac{p_{i,t+1} - p_{i,t}}{p_{i,t}} \, ,
\end{align}
which is what we wish to predict. 
For financial practitioners, it is essential to predict stock returns, since they usually build trading strategies based on predicted future returns.
In academic literature, it has been a central problem to investigate the predictability of stock returns and to construct prediction models which satisfies empirical characteristics, with the hope to reveal market structures.

There are mainly two distinct approaches to predict stock returns: one is that by focusing on a specific stock, we use time series analysis to predict its return, and the other is that we predict cross-sectional relative stock performance for whole stock universe at each time, employing each firm's features.

In this work, we adopt the latter cross-sectional approach, in which we leverage information of firms.
Suppose we have $n$ features for $i$-th stock at time $t$. Such features are compiled to $n$-dimensional vector $X_{i,t}$, by means of which we describe the general formula of our prediction model as follows:
\begin{align} \label{eq:genaral_eq_for_model}
    r_{i,t+1} = F(X_{i,t}) + \epsilon_{i,t}  \, , 
\end{align}
where the form of $F$ is not specified here and will be determined by our choice of models. $\epsilon_{i,t}$ represents an error term.

As for a choice of features, what should explain stock returns is a long-standing subject to study in financial literature and has industriously been investigated.
The celebrated work by Fama and French \cite{eugene1992cross} proposes and empirically shows that returns of individual firms can be explained by the following three factors: market (how the whole market moves), size (how large the market capitalization of stocks is), and value (how the stock price is overvalued or undervalued). Ever since their publication, successive studies have followed in order to find other unknown factors to explain returns, with the result that the number of proposed factors has surpassed a hundred. 
Other than the famous three factors mentioned above, typical factors considered so far include momentum (how big the past return of stocks is) and quality (how stable earnings stocks have).

Regarding the prediction model $F$, linear regression has been traditionally used both for academicians and practitioners, because of its simplicity and interpretability. In this case, \eqref{eq:genaral_eq_for_model} becomes
\begin{align}
    r_{i,t+1} = \sum_{k=1}^{n} X_{i,t,k} \cdot \theta_k + \epsilon_{i,t} \, , 
\end{align}
where $\theta$ is an $n$-dimensional vector of model parameters.
Note that the index $k$ represents $k$-th element of an $n$-dimensional vector. 
In our analysis, we employ this linear regression model as a benchmark. The parameter $\theta$ is determined by the usual ordinary least square regression.

Traditional linear regression models neglect interaction terms between features and nonlinear terms.
Machine learning models shed light on these issues, as is widely reported in, e.g., \cite{abe2018deep,chinco2019sparse,dixon2020deep,10.1093/rfs/hhaa009,gu2021autoencoder}.
In our analysis, we use the neural network models as classical machine learning ones. 
As quantum and quantum-inspired machine learning models, we propose to employ quantum circuit learning and tensor network in return predictions.
The following subsections are devoted to describing these methods and how they can be applied for stock return predictions.

\subsection{Neural network}
We consider a feed-forward neural network, which consists of $L$ layers of affine maps and activation functions. It is formally written as 
\begin{align}
    F^{\rm NN} = \mathsf{W_L} \circ \sigma_{L-1} \circ \cdots \sigma_1 \circ \mathsf{W_1} \, , 
\end{align}
The affine map $W_l$ acts on $n_l$-dimensional input vector $Z_l$ as follows:
\begin{align}
    \mathsf{W}_l(Z_l) = W_l Z_l + b_l \, , 
\end{align}
where $W_l \in \mathbb{R}^{n_{l+1} \times n_l}$ denotes a weight matrix and $b_l \in \mathbb{R}^{n_l}$ a bias vector. 
The activation function $\sigma_l$ is a key to generate a non-linear effect on the model.
Though there are plenty of possibilities for what activation function to use, in our analysis we use the same function $\mathsf{ReLU}$ for all $l = 1, \cdots L$ defined as follows:
\begin{align}
    \sigma_l(x) = \mathsf{ReLU}(x) \equiv \max \{x,0\} \, .
\end{align}

Having these in our hands, we construct return prediction such that 
\begin{align}
    r_{i,t+1} = F^{\rm NN} (X_{i,t}) + \epsilon_{i,t} \, .
\end{align}

\subsection{Quantum circuit learning}\label{sec:qcl}
Among various quantum machine learning algorithms that have been developed recently \cite{cerezo2021varational},
we employ the framework called quantum circuit learning \cite{mitarai2018quantum} in this work.
It is one of the variational quantum algorithms, aiming at application for supervised machine learning problems.
Quantum circuit learning can be regarded as a quantum counterpart of the neural network, since both algorithms try to optimize parameters variationally so that an objective function is minimized. For this reason, quantum circuit learning and similar approaches are also sometimes referred to as quantum neural network \cite{cerezo2021varational}.

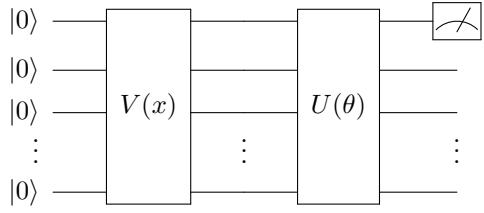
\begin{figure}[tb]
    \centering
    \[
    \Qcircuit @C=2em @R=.7em {
      &&& \lstick{\ket{0}}     & \multigate{5}{V( x)} & \qw    & \multigate{5}{U(\theta)} & \meter  \\
      &&& \lstick{\ket{0}}     & \ghost{V( x)}        & \qw    & \ghost{U(\theta)} &\qw \\
      &&& \lstick{\ket{0}}     & \ghost{V( x)}        & \qw    & \ghost{U(\theta)} &\qw  \\
      &&& \lstick{\vdots}      &                               & \vdots &   & \vdots\\
      &&&&&&& \\
      &&& \lstick{\ket{0}}     & \ghost{V( x)}        & \qw    & \ghost{U(\theta)} & \qw
    }
    \]
    \caption{The general structure of quantum circuit learning, where we have two quantum circuit architectures: Data encoding circuit $V(x)$ and parameterized quantum circuit $U(\theta)$}
    \label{fig:qcl-general}
\end{figure}

Quantum circuit learning consists of the following procedures.
Suppose we have a dataset consisting of input data $\{x_i \}_{i=1}^N$, and corresponding teacher data $\{y_i \}_{i=1}^N$.
First, we construct a quantum circuit $V(x)$ from $x$.
We apply it to some initial state $\ket{\psi_0}$ in order to encode the information of input variables into the quantum state: $\ket{\psi_{in}} = V(x) \ket{\psi_0}$.
Then we prepare a parameterized quantum circuit $U(\theta)$ and apply it to the above state: $\ket{\psi_{out}} = U(\theta) \ket{\psi_{in}}$.
Finally, we measure the expectation value $\bra{\psi_{out}} O \ket{\psi_{out}}$ of some observable $O$.
In this work, we take the Pauli $Z$ operator acting on the first qubit, $Z_1$, as the observable $O$.
It is taken as an output of the algorithm $F^{\mathrm{QCL}}(x,\theta)$. 
The objective function built from $y_i$ and $F^{\mathrm{QCL}}(x,\theta) = \bra{\psi_{out}} Z_1 \ket{\psi_{out}}$ is minimized by varying the parameter $\theta$. 
With the optimized parameter $\theta = \theta^*$, the trained model is given as $F^{\mathrm{QCL}}(x,\theta^*)$. \figref{fig:qcl-general} shows the general circuit of the quantum circuit learning algorithm.

We next explain the construction of quantum circuits for our analysis.
It follows \cite{mitarai2018quantum}.
Remark that we can choose different forms of encoding and parameterized circuits, which may result in different predicting performance. 
We do not touch upon the effects of employing different quantum circuits in this paper, leaving it for future investigation.
The initial state $\ket{\psi_{in}}$ is prepared as $\ket{0}^{\otimes n} $, where we assume the dimension of vectors $x_i$ of input data is $n$.
The encoding circuit $V(x_i)$ is given by 
\begin{align}
    V(x_{i}) = \prod_{j=1}^{n} R_j^Z (\cos^{-1} x_{i,j}^2) R_j^Y (\sin^{-1} x_{i,j}) \, , 
\end{align}
where $R_j^Z$ and $R_j^Y$ represent rotation gates acting on $j$-th qubit:
\begin{align}
    R_j^Z(\phi) = e^{i \phi Z_j / 2}\, , \quad R_j^Y(\phi) = e^{i\phi Y_j / 2} \, .
\end{align}
Note that the input vector $x_i$ must be normalized in the range of $[ -1, 1]$.

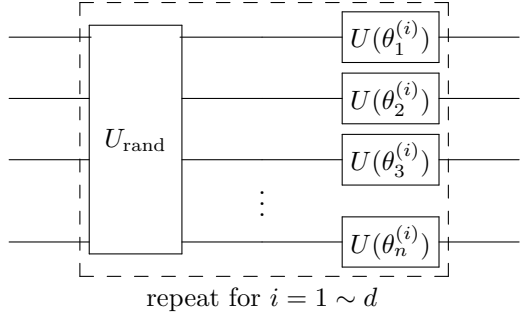
\begin{figure}[tb]
    \centering
    \[
\Qcircuit @C=3em @R=.4em {
	& \multigate{5}{U_{\rm rand}} & \qw&\gate{ U(\theta_1^{(i)}) }    & \qw
	\\
	& \ghost{U_{rand}}   & \qw& \gate{ U(\theta_2^{(i)}) }   & \qw \\
	& \ghost{U_{rand}}   & \qw   & \gate{ U(\theta_3^{(i)}) }   & \qw \\ 
	&  &  \vdots                  &        &  \\ 
	& & & & \\
	& \ghost{U_{rand}} &\qw     & \gate{ U(\theta_n^{(i)}) }   & \qw
	\gategroup{1}{2}{6}{4}{.7em}{--} \\
	&  & & & \\ 
	&  & & & \\
	& & \mbox{repeat for $i=1 \sim d$} & & 
}
\]
    \caption{Our choice of a parameterized quantum circuit in the quantum circuit learning algorithm}
    \label{fig:param-gates}
\end{figure}

Then, our parameterized quantum circuit is constructed as follows 
\begin{align}\label{eq:ansatz}
    U(\theta) = \prod_{i=1}^{d}  \left( \prod_{j=1}^n U(\theta_j^{(i)})  U_{\rm rand}\right) \, ,
\end{align}
which is illustrated in \figref{fig:param-gates}.
Here, $U_{\rm rand}$ denotes a time evolution gate for the following Hamiltonian:
\begin{align}
    U_{\rm rand} = e^{-i H\tau} , \quad H = \sum_{j=1}^n a_j X_j + \sum_{j=1}^n \sum_{k=1}^{j-1} J_{jk} Z_j Z_k \, , 
\end{align}
where $a_j$ and $J_{j,k}$ are randomly taken from a uniform distribution on $[-1,1]$ and $\tau$ represents a time length of the evolution. Both of these parameters are fixed during the algorithm.
$U(\theta_j^{(i)})$ denotes a sequence of rotation gates on $j$-th qubit:
\begin{align}
    U(\theta_j^{(i)}) = R_j^X\left(\theta_{j1}^{(i) }  \right)  R_j^Z\left(\theta_{j2}^{(i) }  \right)  R_j^X \left(\theta_{j3}^{(i) }  \right) \, .
\end{align}
where $R_j^X(\phi) = e^{i \phi X_j / 2}$.
$U_{\rm rand}$ and $U(\theta_j^{(i)})$ are repeatedly applied to the state for $d$ times, resulting in the whole gate $U(\theta) $ in Eq. \eqref{eq:ansatz}.

Equipped with these gates, quantum circuit learning can be used in return prediction such that
\begin{align}
    r_{i,t+1} = F^{\rm QCL}(X_{i,t}, \theta) + \epsilon_{i,t} \, .
\end{align}

\subsection{Tensor Network}
Tensor network enables us to obtain effective representations of quantum wavefunctions that live on exponentially large dimensional Hilbert space.
This is beneficial not only for quantum physics but also for machine learning problems, as tensor network enables us to manipulate a high-dimensional feature space.

The matrix product state (MPS), one of the best-studied and understood tensor networks among all types of ones, 
is employed in our analysis.
MPS is defined as follows.
Suppose we have an $n$-th order tensor $T$, whose component is given by $T_{i_1 \cdots i_n}$.  
The MPS is a representation of such a tensor $T$ by a product of smaller tensors:
\begin{align}
    T_{i_1 \cdots i_n} = \sum_{\alpha_1 \cdots \alpha_n} A^{(1)}_{i_1 \alpha_1}  A^{(2)}_{i_2 \alpha_1 \alpha_2} \cdots  A^{(n)}_{i_n, \alpha_n } \, ,
\end{align}
where the range of indices $\alpha_i$ is called the bond dimension $m$. 

We follow the approach taken in \cite{novikov2016exponential,efthymiou2019tensornetwork,stoudenmire2018learning} to apply the MPS to our purposes.
Consider input vector $x$ and a feature map $\Phi(x)$ which maps $x$ to an $n$-th order tensor defined as
\begin{align}
    \Phi_{i_1 \cdots i_n}(x) = \phi_{i_1}(x_1) \phi_{i_2}(x_2)  \cdots \phi_{i_n}(x_n) \, ,
\end{align}
where
\begin{align}
    \phi (x_j) = \left(
        \begin{array}{c}
        1 \\
        x_j \\
        \end{array}
        \right) \, .
\end{align}
We construct a model regression function with an MPS $W^{\rm MPS}$ which acts as variational parameters to be trained as
\begin{align}
    y = F^{\rm MPS}(x,W) =  \sum_{i_1\cdots i_n} W_{i_1\cdots i_n}^{\rm MPS} \Phi_{i_1\cdots i_n}(x) \, .
\end{align}
We use this function $F^{\rm MPS}(W,x) $ in return prediction:
\begin{align}
    r_{i,t+1} = F^{\rm MPS}(X_{i,t},W) + \epsilon_{i,t} \, .
\end{align}

\subsection{Optimization procedure}
Now that we have introduced both classical and quantum machine learning models we test in our analysis, let us briefly describe how the training of models is performed.
In this subsection, we denote all the prediction models as $F(X_{i,t}, \theta)$ where $\theta$ represent parameters for corresponding model, unless otherwise noted.
Given true return data $r_{i,t}$ and predicted one $\tilde{r}_{i,t} = F(X_{i,t})$, our objective function $E$ to be minimized is the mean squared error:
\begin{align}
    E = \frac{1}{NT} \sum_{t=1}^T \sum_{i=1}^{N} ( \tilde{r}_{i,t} - r_{i,t} )^2 \, .
\end{align}
To achieve the minimum, we utilize the stochastic gradient descent technique for all models, which is a common prescription in learning of neural networks. 
In this framework, parameters are subsequently updated such as 
\begin{align}
    \theta \leftarrow \theta - \eta  \nabla_\theta E \, ,
\end{align}
where $\eta$ represents a hyperparameter and the explicit formula for updating parameters depends on the type of optimizers.
As for the quantum circuit learning model $F = F^{\rm QCL}$, the gradient is calculated by the so-called parameter-shift rule \cite{mitarai2018quantum,schuld2019evaluating}.

It is worth mentioning that, in tensor network, gradient descent technique is not a standard way to optimize parameters, since a more physics-oriented optimization algorithm called density matrix renormalization group (DMRG) \cite{white1992density} prevails in many physics applications and is also used in machine learning one \cite{stoudenmire2016supervised}.  
We, however, work with gradient descent in our analysis, as it is simple to implement on high-level API such as TensorFlow \cite{tensorflow2015-whitepaper} and allows us to compare with other models on equal footing.
Note that the DMRG approach is thought to be more sophisticated in updating parameters than gradient descent; therefore, it would be interesting to investigate the difference of performances in optimizing tensor network models. See \cite{efthymiou2019tensornetwork} for more details.

\section{Experiment}
\label{sec:experiment}
In this section, we show our empirical study to evaluate how our proposed models perform in return prediction. 
Our criteria for the evaluation is how profitable our models are, which can be measured by applying models in investment strategies.
For this purpose, we adopt an investment strategy based on models' predictions and conduct the backtesting experiment on past historical data.
In the following, we explain our dataset and methodology of the investment strategy, then discuss results of backtesting.

\subsection{Dataset}
Our dataset, or investment universe $U_t$, is a set of the Japanese stocks that are constituents of TOPIX500 index.
TOPIX500 is a Japanese stock market index, consisting of the 500 most liquid stocks with the largest values of market capitalization among members of stocks listed on the Tokyo Stock Exchange.

Input features we use are summarized in \tblref{tab:features}.
We consider 10 features, which is rather a small number compared to general machine learning models for stock return predictions, where we typically employ as many as tens to hundreds of features to gain expressibility and accuracy.
This is due to the fact that our quantum circuit learning architecture requires one qubit for each feature; the more qubits we use, the more computationally intense the simulation of quantum circuits becomes. 
We therefore limit the number of features to $n=10$ so that our backtesting experiment can be conducted within reasonable computational time.

As a preprocessing, all features and returns are cross-sectionally ranked at each time step \cite{10.1093/rfs/hhaa009,nakagawa2020ric}: the $i$th feature of the $l$th stock at time $t$, $x_{i,t,l}$, is converted to $(\rho_{i,t,l})/(N_t-1)$, where $\rho_{i,t,l}$ is the rank of $x_{i,t,l}$ among $\{x_{i,t,l}\}_{i=1,...,N_t}$ in the ascending order. 
\begin{table}
    \centering
    \caption{The list of features and their descriptions}
    \label{tab:features}
    \begin{tabular}{@{}lll@{}} \toprule 
    Factor                    & Feature                   & Description                                                                                                                  \\ 
    \midrule
    \multirow{3}{*}{Value}    & Book-Value to Price Ratio & Net Asset / Market Value                                                                                                     \\ 
                              & Earning to Price Ratio    & Net Profit / Market Value                                                                                                    \\ 
                              & Sales to Price Ratio      & Sales / Market Value                                                                                                         \\ 
    \hline
    Quality                   & Return on Equity          & Net Profit / Net Asset                                                                                                       \\ 
    \hline
    \multirow{4}{*}{Momentum} & Momentum (1-month)        & Stock Returns in the last month                                                                                              \\ 
                              & Momentum (3-month)        & Stock Returns in the past 3 months                                                                                           \\ 
                              & Momentum (6-month)        & Stock Returns in the past 6months                                                                                            \\ 
                              & Momentum (12-month)       & Stock Returns in the past 12 months                                                                                          \\ 
    \hline
    Size                      & Market Capitalization     & log(Market Value)                                                                                                            \\ 
    \hline
    Market                    & Beta                      & \begin{tabular}[c]{@{}l@{}}Regression coefficient of stock returns \\ and market return (TOPIX) over 60 months\end{tabular}  \\
    \hline
    \end{tabular}
    \end{table}

\subsection{Investment strategy}
The investment strategy that we take in this work is as follows. 
Our backtesting period goes from June 2008 to May 2021, during which we make investment decisions on a monthly basis and let $t$ denote the end of each month. Subsequently, our subject to predict is then a one-month future return.

\begin{figure}
    \centering
    \includegraphics[width=90mm]{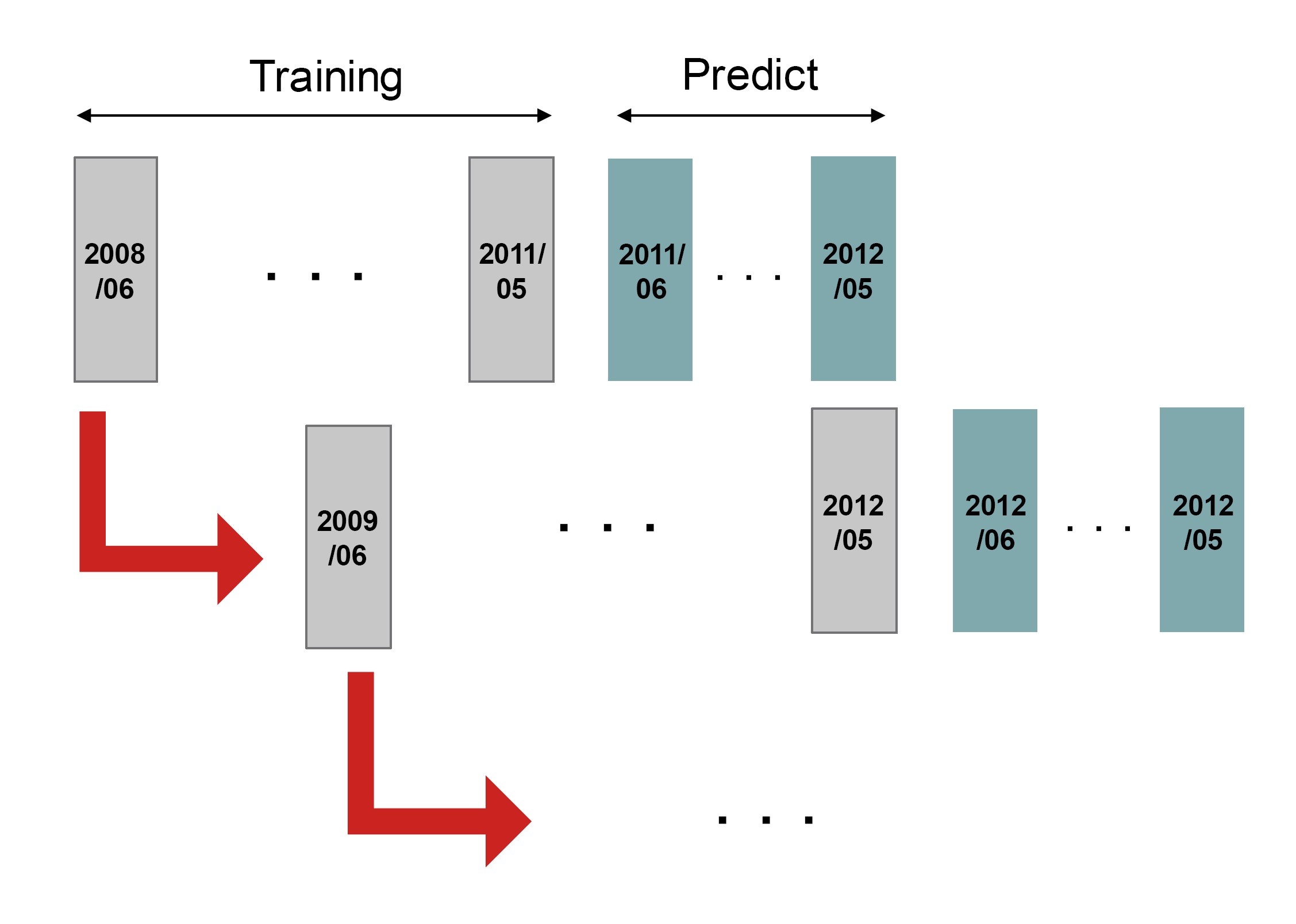}
    \caption{The concept of our backtesting experiment, showing that we take three years as a training period and subsequent one year as a test period, rolling this process until the end of the backtesting period}
    \label{fig:WF}
\end{figure}

At the beginning of backtesting, we take three-year samples (June 2008 - May 2011) as the training dataset to train the model, and the following one-year ones (June 2011 - May 2012) as the test dataset to predict returns.
We then roll this procedure forward until the end of the backtesting period. See \figref{fig:WF} for its design.
In short, we repeatedly make a prediction for the forthcoming year from the most recent three-year samples, but only re-estimate models once a year, not every month, in order to avoid computationally intensive estimation, which is a severe problem for quantum circuit learning running on a simulator.

At each time step $t$ we sort stocks in descending order based on predicted returns $\tilde{r}_{i,t+1}$ and define a set of stocks belonging to the top quintile as $H_t$. Assuming our models correctly predict stock returns, $H_t$ should represent most profitable stocks among the whole universe $U_t$. 
On that account, we go long, or buy, these stocks with equal weight. 
The portfolio return between $t$ and $t+1$ is then given by 
\begin{align}
    r_{\mathrm{port},t+1} = \frac{1}{|H_t|} \sum_{i \in H_t} r_{i,t+1} \, , 
\end{align}
where $|H_t|$ denotes the number of stocks in $H_t$.
We repeat this process and measure the portfolio performance over the backtesting period.

To test the performance of our investment strategy, the common approach is to set up a benchmark portfolio and evaluate excess return between our portfolio and the benchmark. 
In this work, we use the TOPIX500 index as a benchmark; therefore, the excess return is defined as 
\begin{align}
    \alpha_t = r_{\mathrm{port},t} - r_{\mathrm{TOPIX500},t} \, , 
\end{align}
where $r_{\mathrm{TOPIX500},t}$ denotes the return of the TOPIX500 index at time $t$.
The metrics of the portfolio performance we employ are the following three quantities, all of which are constructed from the time series of $\alpha_t$:
\begin{align}
    \mathrm{ER} &= \prod_{t=1}^T (1 + \alpha_t)^{12/T} - 1 \, , \\
    \mathrm{TE} &= \sqrt{\frac{12}{T-1}\sum_t^T \left(\alpha_t - \bar{\alpha}_t)^2 \right)}  \, , \\
    \mathrm{IR} &= \mathrm{ER} / \mathrm{TE} \, , 
\end{align}
with $\bar{\alpha}_t = 1/T \sum_{t=1}^T \alpha_t $.
Here, ER represents an annualized excess return, TE (tracking error) denotes the corresponding standard deviation, and IR is the so-called information ratio, which expresses the risk-adjusted excess return of the portfolio.

\subsection{Model architectures}
We summarize the detailed setting of our models.
As a traditional model, we use the linear regression which we denote \textsf{Linear}.
In all models we consider except for the linear regression, the number of parameters is set to be in the same order for fair comparison. 
We use Adam optimizer in the training, where the number of epochs is also fixed to 20 in all machine learning models.

\paragraph*{Neural Network}
We prepare two distinct neural network models, which differ in the number of hidden layers.
\begin{itemize}
    \item \textsf{NN1} denotes the neural network model with $L=3$ layers whose nodes are given by $(10,7,1)$. This model has 92 parameters to be trained.
    \item \textsf{NN2} denotes the neural network model with $L=4$ layers whose nodes are given by $(10,5,4,1)$. This model has 93 parameters to be trained.
\end{itemize}
As mentioned earlier, we stick to use the ReLU function as the activation function. TensorFlow \cite{tensorflow2015-whitepaper} is used to implement the model.

\paragraph*{Quantum Circuit Learning}
We denote our quantum circuit learning model by \textsf{QCL}. The number of qubits is 10, which is the same as the number of input features. The depth of parameterized gates is set to $d=3$. The number of parameters is consequently 90. 
We use a quantum circuit simulator Qulacs \cite{suzuki2020qulacs} to implement and simulate quantum circuits.
We have conducted the numerical experiments in a noiseless setting.

\paragraph*{Tensor Network}
We denote our tensor network model by \textsf{TN}.
We set the bond dimension to $m=2$.
The number of parameters is then 76 in this setting.
We use TensorNetwork \cite{roberts2019tensornetwork} as well as TensorFlow for its implementation.

\subsection{Backtesting result}

\begin{figure}[H]
    \centering
    \includegraphics[width=85mm]{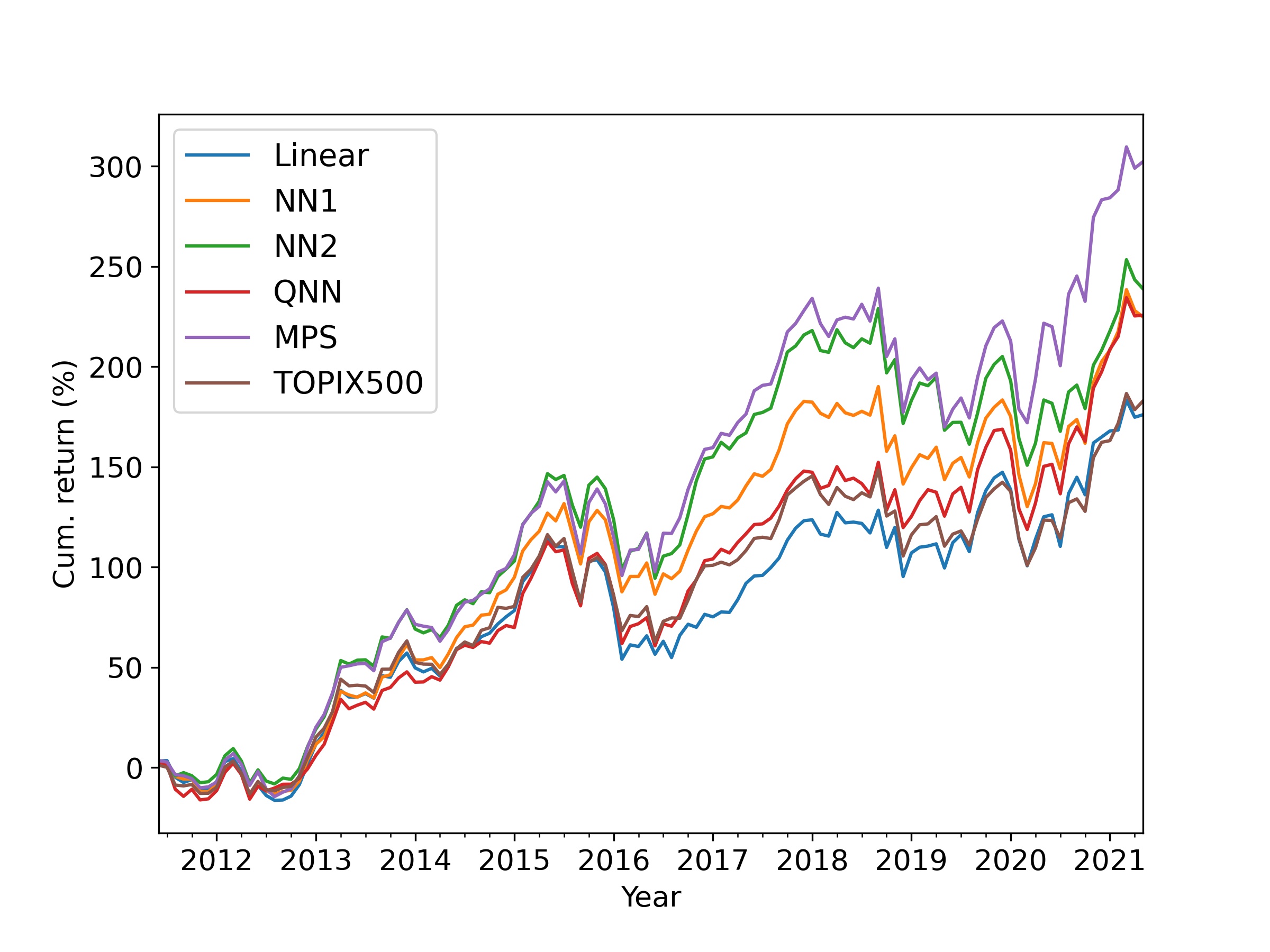}
    \caption{The cumulative returns of portfolios constructed by various methods and that of TOPIX500}
    \label{fig:cum_ret}
\end{figure}

\begin{figure}[H]
    \centering
    \includegraphics[width=85mm]{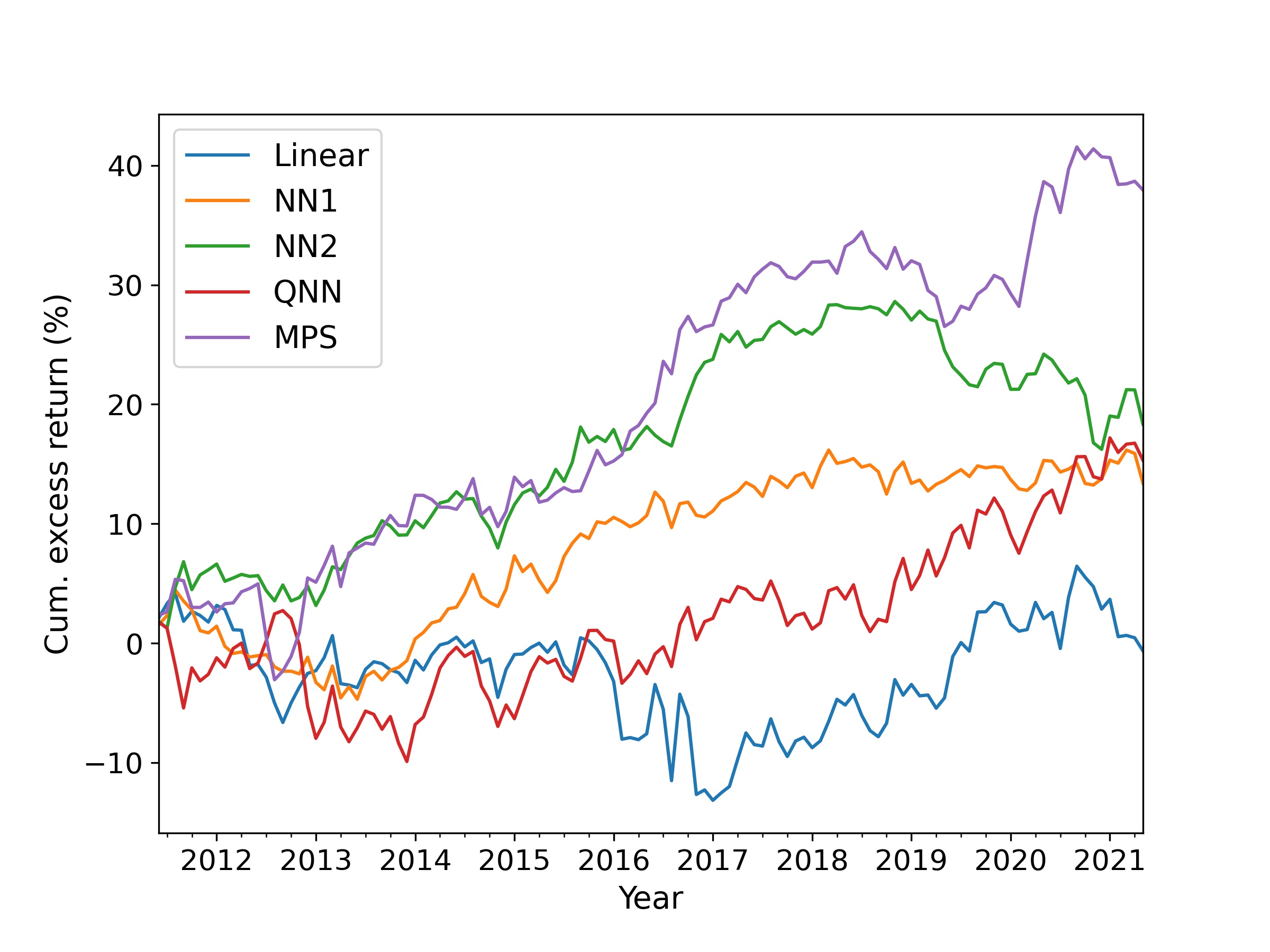}
    
    \caption{The cumulative excess returns of portfolios constructed by various methods over TOPIX500}
    \label{fig:excess_ret}
\end{figure}

\tblref{tab:result} summarizes the results of our empirical backtesting.
See also \figref{fig:cum_ret} for cumulative returns of portfolios and \figref{fig:excess_ret} for cumulative excess returns.
We observe that the tensor network model \textsf{TN} has the best performance in regard to both the excess return and the information ratio.
On the other hand, the quantum circuit learning model \textsf{QCL} has competitive performance with the neural network model with respect to the excess return; however, it has a larger value of $\mathrm{TE}$, which in turn results in inferior risk-adjusted return $\mathrm{IR}$.
\begin{table}[h] 

    \centering
    \caption{The empirical result of backtesting in TOPIX500 universe (Bold characters show the best numbers in each metrics)}
    \begin{tabular}{@{}llllll@{}}
        \toprule
         & \textsf{Linear} & \textsf{NN1} & \textsf{NN2} & \textsf{QCL} & \textsf{TN} \\ 
        \midrule
        ER (\%) & -0.28 & 1.27 & 1.76 & 1.35 & \textbf{3.71} \\
        
        TE (\%) & 6.64 & \textbf{3.79} & 4.28 & 6.18 & 5.41 \\ 
        
        IR & -0.04 & 0.34 & 0.41 & 0.22 & \textbf{0.69}        \\
        \botrule
    \end{tabular}

    \label{tab:result}
\end{table}

From \figref{fig:excess_ret}, before 2016, \textsf{QCL} has the approximately same performance as \textsf{Linear}. This implies that \textsf{QCL} at least learns the linear relationship between input features as is expected. After 2016, on the other hand, \textsf{QCL} continues to outperform  \textsf{Linear}, which might be because  \textsf{QCL} is able to learn non-linear relationships as well. 
What is more, in these recent market environments, \textsf{QCL} can successfully predict stock returns and gain the excess returns, beating classical models. See Appendix \ref{sec:result_after2016} for numerics and graphs. 
We also find that during the last three years in the backtesting period, neural network models perform poorly.
It suggests that neural networks used in this analysis tend to overfit to the previous market environment and fail in adapting to the latest one.

The tensor network model \textsf{TN} has the best performance over other models in spite of the lowest number of parameters.
It illustrates that \textsf{TN} can possibly have effective architectures to learn financial data, to say nothing of the possibility to capture non-linearity among features.
It should be further investigated in the future whether this superiority holds when we increase the number of features and parameters in models.

\section{Conclusion and discussion}
\label{sec:conclusion}
In this paper, we propose to use quantum and its inspired algorithms to predict stock returns. We especially test quantum circuit learning and tensor network as the proposed model against the classical models, namely linear regression and neural network. 
In order to evaluate their capabilities, we consider the investment strategy based on predicted returns by classical and quantum models. 
We then conduct backtesting over 10 years in the Japanese stock market. 

Our finding is that the tensor network model outperforms classical models, while the quantum circuit learning model archives comparable performance with the neural network models but with higher risk.
As is expected, both proposed models seem to learn non-linear relationships between input features, implied by their superior performance against linear regression.
Although the performance of the neural network models deteriorates in the latest years, our proposed models successfully continue to gain the excess return. 
These differences in the performance can be related to the overfitting problem in machine learning and market instability in these periods. 
We therefore speculate that quantum techniques can have a good control of the overfitting problem, which is originally suggested in \cite{mitarai2018quantum}. 
It is, however, unclear whether the hypothesis is true; further examination on this issue should be conducted.

Lastly, we comment on several open problems for future exploration.
\begin{itemize}
    \item In this work, we evaluate models' capabilities in the Japanese stock market. It should be examined if quantum models work in other countries, e.g., the United States, or in the global market. \cite{nakagawa2020ric} studies the transfer learning of neural network in the investment problem between various markets. Whether transfer learning in the quantum model is also effective or not is another interesting research direction. 
    \item While we study the predictability of stocks, it would be interesting whether quantum machine learning is applicable for other assets, such as bonds or currencies. See \cite{published_papers/35761356,Poh89} for the machine learning approach in these assets.
    \item As is explained in Section \ref{sec:method}, there are two approaches towards the return prediction, one of which is the cross-section prediction we employ. The other way, namely the time-series approach, can be applied in quantum machine learning. In classical neural networks, recurrent neural network and its variants are developed and widely investigated in financial literature \cite{bao2017deep,kim2019enhancing,lim2019enhancing,duan2021learning}.
    It would be interesting to apply the quantum counterpart of such recurrent networks in financial analysis. See \cite{takaki2021learning,bausch2020recurrent} for the existing literature of quantum recurrent neural networks.
\end{itemize}



\section*{Data Availability}
The datasets analyzed during the current study are not publicly available due to the licensing agreements with the data providers.

\section*{Statements and Declarations}
\subsection*{Competing Interests}
The authors declare no competing interests.

\subsection*{Authors' contributions}
All authors contributed to the study conception and design. Data collection and analysis were performed by N.K. The first draft of the manuscript was written by N.K and all authors commented on previous versions of the manuscript. All authors read and approved the final manuscript.

\section*{Acknowledgement}

This work was partially supported by
KAKENHI Grant Number JP22K11924 from JSPS, MEXT Q-LEAP Grant No. JPMXS0120319794, and JST COI-NEXT No. JPMJPF2014.

\begin{appendices}
    \section{Backtesting result from 2016}
    \label{sec:result_after2016}
\begin{figure}[H]
        \centering
        \includegraphics[width=85mm]{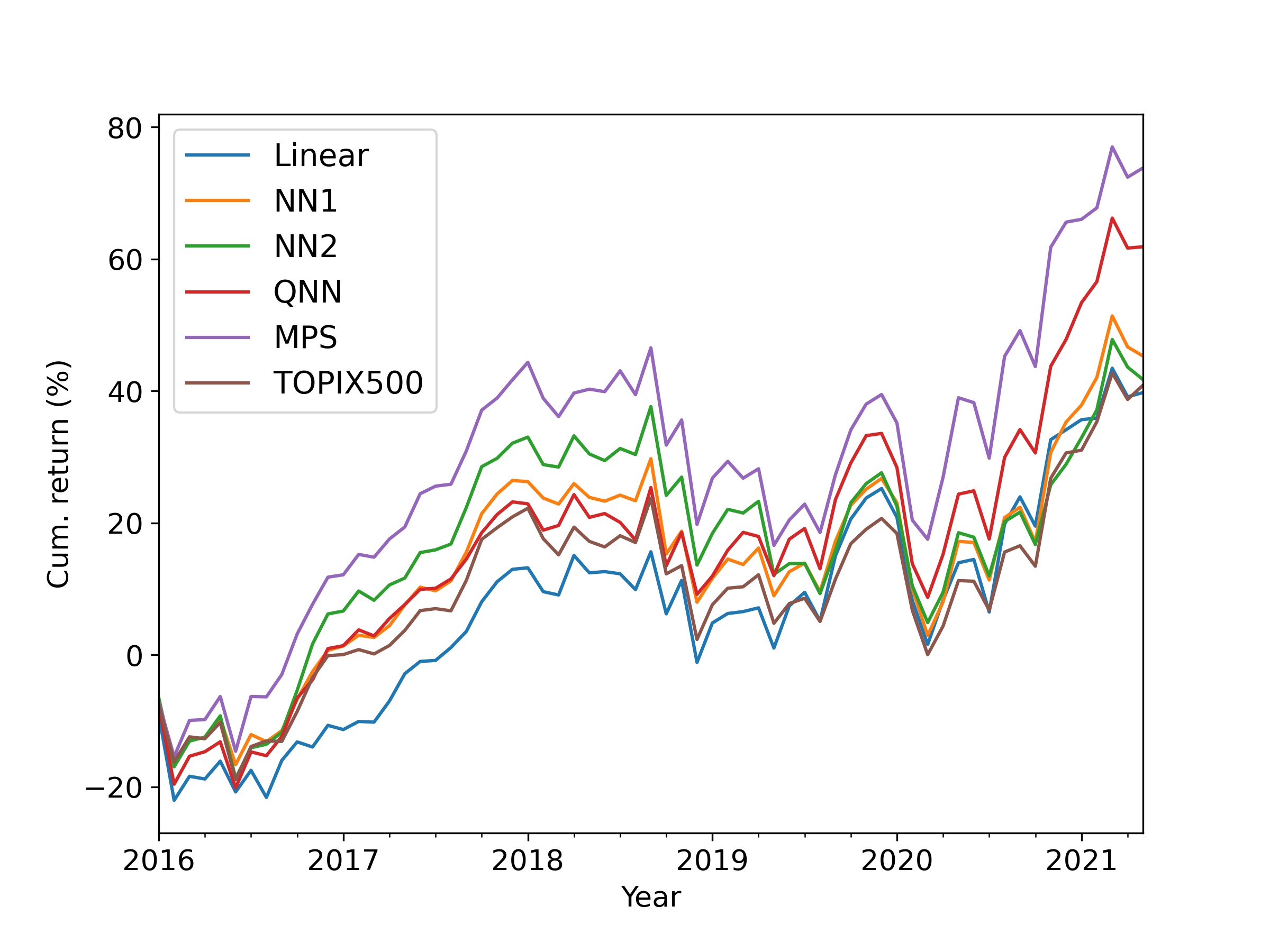}
        \caption{The cumulative returns of portfolios constructed by various methods from 2016 and that of TOPIX500}
        \label{fig:cum_ret_from2016}
    \end{figure}

\begin{figure}[H]
        \centering
        \includegraphics[width=85mm]{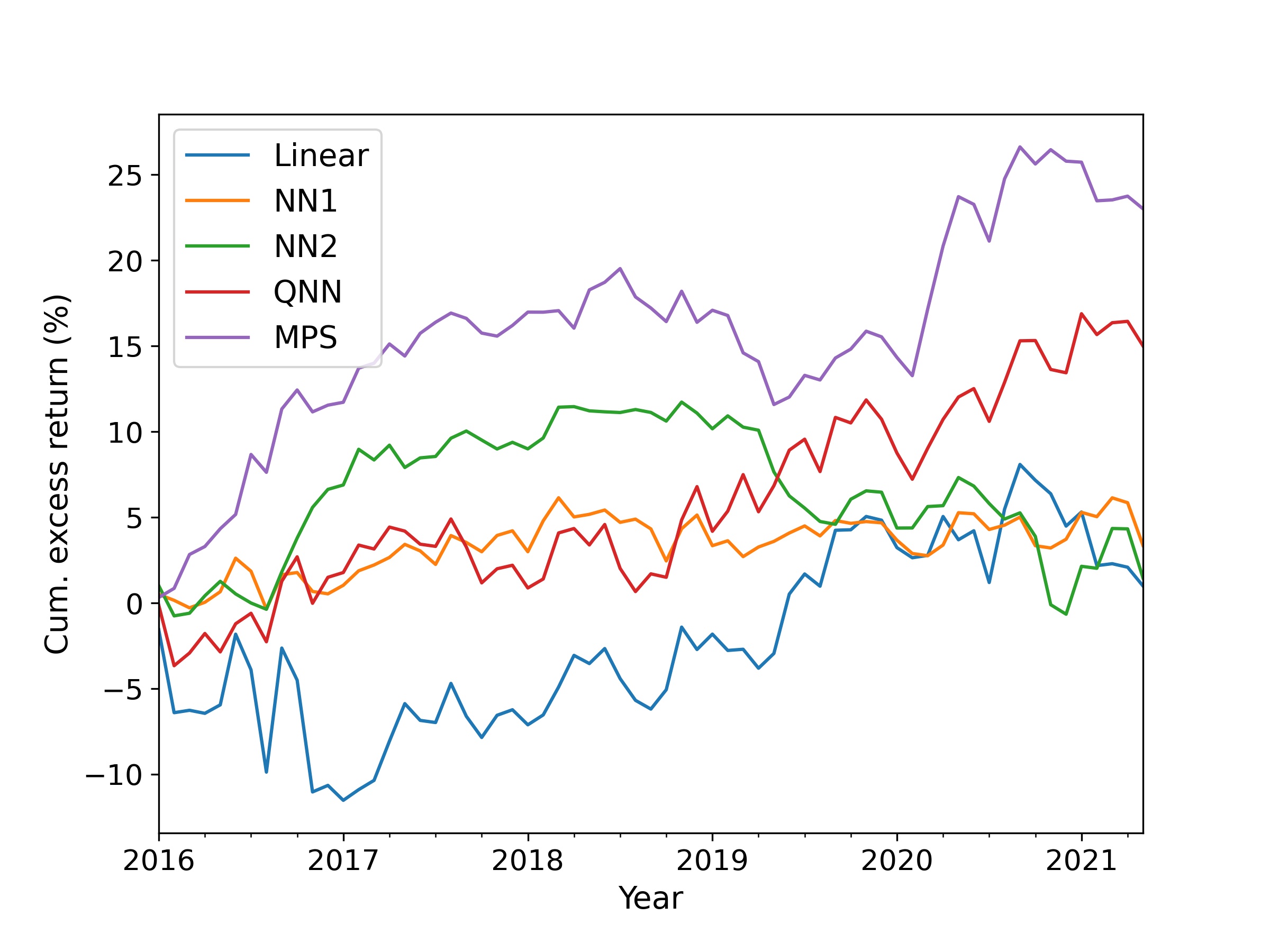}
        
        \caption{The cumulative excess returns of portfolios constructed by various methods over TOPIX500 from 2016}
        \label{fig:excess_ret_from2016}
    \end{figure}

    In this Appendix, we show the backtesting results only from 2016 in \tblref{tab:result_from2016}, \figref{fig:cum_ret_from2016} and \figref{fig:excess_ret_from2016}.

    \begin{table}[h] 
    
        \centering
        \caption{The empirical result of backtesting in TOPIX500 universe (Bold characters show the best numbers in each metrics)}
        \begin{tabular}{@{}llllll@{}}
            \toprule
             & \textsf{Linear} & \textsf{NN1} & \textsf{NN2} & \textsf{QCL} & \textsf{TN} \\ 
            \midrule
            ER (\%) & -0.12 & 0.56 & 0.18 & 2.63 & \textbf{4.20} \\
        
            TE (\%) & 7.82 & \textbf{3.48} & 4.31 & 5.85 & 5.14 \\ 
    
            IR & -0.02 & 0.16 & 0.04 & 0.45 & \textbf{0.82}        \\
            \botrule
        \end{tabular}

        \label{tab:result_from2016}
    \end{table}
    
    \end{appendices}

\bibliography{ref}

\begin{thebibliography}{31}
\providecommand{\natexlab}[1]{#1}
\providecommand{\url}[1]{{#1}}
\providecommand{\urlprefix}{URL }
\providecommand{\doi}[1]{\url{https://doi.org/#1}}
\providecommand{\eprint}[2][]{\url{#2}}
 \bibcommenthead

\bibitem[{Abadi et~al(2015)Abadi, Agarwal, Barham, Brevdo, Chen, Citro,
  Corrado, Davis, Dean, Devin, Ghemawat, Goodfellow, Harp, Irving, Isard, Jia,
  Jozefowicz, Kaiser, Kudlur, Levenberg, Man\'{e}, Monga, Moore, Murray, Olah,
  Schuster, Shlens, Steiner, Sutskever, Talwar, Tucker, Vanhoucke, Vasudevan,
  Vi\'{e}gas, Vinyals, Warden, Wattenberg, Wicke, Yu, and
  Zheng}]{tensorflow2015-whitepaper}
Abadi M, Agarwal A, Barham P, et~al (2015) {TensorFlow}: Large-scale machine
  learning on heterogeneous systems.
  \urlprefix\url{https://www.tensorflow.org/}, software available from
  tensorflow.org

\bibitem[{Abe and Nakayama(2018)}]{abe2018deep}
Abe M, Nakayama H (2018) Deep learning for forecasting stock returns in the
  cross-section. In: Pacific-Asia conference on knowledge discovery and data
  mining, Springer, pp 273--284

\bibitem[{Arute et~al(2019)Arute, Arya, Babbush, Bacon, Bardin, Barends,
  Biswas, Boixo, Brandao, Buell et~al}]{arute2019quantum}
Arute F, Arya K, Babbush R, et~al (2019) Quantum supremacy using a programmable
  superconducting processor. Nature 574(7779):505--510

\bibitem[{Bao et~al(2017)Bao, Yue, and Rao}]{bao2017deep}
Bao W, Yue J, Rao Y (2017) A deep learning framework for financial time series
  using stacked autoencoders and long-short term memory. PloS one
  12(7):e0180944

\bibitem[{Bausch(2020)}]{bausch2020recurrent}
Bausch J (2020) Recurrent quantum neural networks. Advances in neural
  information processing systems 33:1368--1379

\bibitem[{Cerezo et~al(2021)Cerezo, Arrasmith, Babbush, Benjamin, Endo, Fujii,
  McClean, Mitarai, Yuan, Cincio, and Coles}]{cerezo2021varational}
Cerezo M, Arrasmith A, Babbush R, et~al (2021) Variational quantum algorithms.
  Nature Reviews Physics 3(9):625--644

\bibitem[{Chinco et~al(2019)Chinco, Clark-Joseph, and Ye}]{chinco2019sparse}
Chinco A, Clark-Joseph AD, Ye M (2019) Sparse signals in the cross-section of
  returns. The Journal of Finance 74(1):449--492

\bibitem[{Dixon and Polson(2020)}]{dixon2020deep}
Dixon M, Polson N (2020) Deep fundamental factor models. SIAM Journal on
  Financial Mathematics 11(3):SC26--SC37

\bibitem[{Duan and Kashima(2021)}]{duan2021learning}
Duan J, Kashima H (2021) Learning to rank for multi-step ahead time-series
  forecasting. IEEE Access 9:49372--49386

\bibitem[{Efthymiou et~al(2019)Efthymiou, Hidary, and
  Leichenauer}]{efthymiou2019tensornetwork}
Efthymiou S, Hidary J, Leichenauer S (2019) Tensornetwork for machine learning.
  arXiv preprint arXiv:190606329

\bibitem[{Eugene and French(1992)}]{eugene1992cross}
Eugene F, French K (1992) The cross-section of expected stock returns. Journal
  of Finance 47(2):427--465

\bibitem[{Fannes et~al(1992)Fannes, Nachtergaele, and
  Werner}]{fannes1992finitely}
Fannes M, Nachtergaele B, Werner RF (1992) Finitely correlated states on
  quantum spin chains. Communications in mathematical physics 144(3):443--490

\bibitem[{Gu et~al(2020)Gu, Kelly, and Xiu}]{10.1093/rfs/hhaa009}
Gu S, Kelly B, Xiu D (2020) {Empirical Asset Pricing via Machine Learning}. The
  Review of Financial Studies 33(5):2223--2273. \doi{10.1093/rfs/hhaa009},
  \urlprefix\url{https://doi.org/10.1093/rfs/hhaa009},
  {\href{https://arxiv.org/abs/https://academic.oup.com/rfs/article-pdf/33/5/2223/33209812/hhaa009.pdf}{{https://academic.oup.com/rfs/article-pdf/33/5/2223/33209812/hhaa009.pdf}}}

\bibitem[{Gu et~al(2021)Gu, Kelly, and Xiu}]{gu2021autoencoder}
Gu S, Kelly B, Xiu D (2021) Autoencoder asset pricing models. Journal of
  Econometrics 222(1):429--450

\bibitem[{Huggins et~al(2019)Huggins, Patil, Mitchell, Whaley, and
  Stoudenmire}]{huggins2019towards}
Huggins W, Patil P, Mitchell B, et~al (2019) Towards quantum machine learning
  with tensor networks. Quantum Science and technology 4(2):024001

\bibitem[{Kim(2019)}]{kim2019enhancing}
Kim S (2019) Enhancing the momentum strategy through deep regression.
  Quantitative Finance 19(7):1121--1133

\bibitem[{Lim et~al(2019)Lim, Zohren, and Roberts}]{lim2019enhancing}
Lim B, Zohren S, Roberts S (2019) Enhancing time-series momentum strategies
  using deep neural networks. The Journal of Financial Data Science 1(4):19--38

\bibitem[{Madsen et~al(2022)Madsen, Laudenbach, Askarani, Rortais, Vincent,
  Bulmer, Miatto, Neuhaus, Helt, Collins et~al}]{madsen2022quantum}
Madsen LS, Laudenbach F, Askarani MF, et~al (2022) Quantum computational
  advantage with a programmable photonic processor. Nature 606(7912):75--81

\bibitem[{Mitarai et~al(2018)Mitarai, Negoro, Kitagawa, and
  Fujii}]{mitarai2018quantum}
Mitarai K, Negoro M, Kitagawa M, et~al (2018) Quantum circuit learning.
  Physical Review A 98(3):032309

\bibitem[{Nakagawa et~al(2020)Nakagawa, Abe, and Komiyama}]{nakagawa2020ric}
Nakagawa K, Abe M, Komiyama J (2020) Ric-nn: a robust transferable deep
  learning framework for cross-sectional investment strategy. In: 2020 IEEE 7th
  International Conference on Data Science and Advanced Analytics (DSAA), IEEE,
  pp 370--379

\bibitem[{Novikov et~al(2016)Novikov, Trofimov, and
  Oseledets}]{novikov2016exponential}
Novikov A, Trofimov M, Oseledets I (2016) Exponential machines. arXiv preprint
  arXiv:160503795

\bibitem[{Poh et~al(2022)Poh, Lim, Zohren, and Roberts}]{Poh89}
Poh D, Lim B, Zohren S, et~al (2022) Enhancing cross-sectional currency
  strategies by context-aware learning to rank with self-attention. The Journal
  of Financial Data Science 4(3):89--107. \doi{10.3905/jfds.2022.1.099},
  \urlprefix\url{https://jfds.pm-research.com/content/4/3/89},
  {\href{https://arxiv.org/abs/https://jfds.pm-research.com/content/4/3/89.full.pdf}{{https://jfds.pm-research.com/content/4/3/89.full.pdf}}}

\bibitem[{Preskill(2018)}]{Preskill2018quantumcomputingin}
Preskill J (2018) Quantum {C}omputing in the {NISQ} era and beyond. {Quantum}
  2:79. \doi{10.22331/q-2018-08-06-79},
  \urlprefix\url{https://doi.org/10.22331/q-2018-08-06-79}

\bibitem[{Roberts et~al(2019)Roberts, Milsted, Ganahl, Zalcman, Fontaine, Zou,
  Hidary, Vidal, and Leichenauer}]{roberts2019tensornetwork}
Roberts C, Milsted A, Ganahl M, et~al (2019) Tensornetwork: A library for
  physics and machine learning. \eprint{1905.01330}

\bibitem[{Schuld et~al(2019)Schuld, Bergholm, Gogolin, Izaac, and
  Killoran}]{schuld2019evaluating}
Schuld M, Bergholm V, Gogolin C, et~al (2019) Evaluating analytic gradients on
  quantum hardware. Physical Review A 99(3):032331

\bibitem[{Stoudenmire and Schwab(2016)}]{stoudenmire2016supervised}
Stoudenmire E, Schwab DJ (2016) Supervised learning with tensor networks.
  Advances in Neural Information Processing Systems 29

\bibitem[{Stoudenmire(2018)}]{stoudenmire2018learning}
Stoudenmire EM (2018) Learning relevant features of data with multi-scale
  tensor networks. Quantum Science and Technology 3(3):034003

\bibitem[{Suimon et~al(2020)Suimon, Sakaji, Izumi, Shimada, and
  Matsushima}]{published_papers/35761356}
Suimon Y, Sakaji H, Izumi K, et~al (2020) Japanese interest rate forecast
  considering the linkage of global markets using machine learning methods.
  International Journal of Smart Computing and Artificial Intelligence
  4(1):1--17. \doi{10.52731/ijscai.v4.i1.500}

\bibitem[{Suzuki et~al(2020)Suzuki, Kawase, Masumura, Hiraga, Nakadai, Chen,
  Nakanishi, Mitarai, Imai, Tamiya et~al}]{suzuki2020qulacs}
Suzuki Y, Kawase Y, Masumura Y, et~al (2020) Qulacs: a fast and versatile
  quantum circuit simulator for research purpose. arXiv preprint
  arXiv:201113524

\bibitem[{Takaki et~al(2021)Takaki, Mitarai, Negoro, Fujii, and
  Kitagawa}]{takaki2021learning}
Takaki Y, Mitarai K, Negoro M, et~al (2021) Learning temporal data with a
  variational quantum recurrent neural network. Physical Review A 103(5):052414

\bibitem[{White(1992)}]{white1992density}
White SR (1992) Density matrix formulation for quantum renormalization groups.
  Physical review letters 69(19):2863

\end{thebibliography}

\end{document}